\documentclass[letterpaper, 10 pt, conference]{ieeeconf}  

\IEEEoverridecommandlockouts                              

\usepackage{amsmath,amsfonts}
\usepackage{graphics} 
\usepackage{epsfig} 
\usepackage{mathptmx} 
\usepackage{times} 
\usepackage{amsmath} 
\usepackage{amssymb}  
\usepackage{pifont}
\usepackage{xcolor}
\usepackage{array}
\usepackage{booktabs}
\usepackage{algorithm}
\usepackage{algpseudocode}
\usepackage{caption}
\usepackage{balance}
\usepackage{textcomp}
\usepackage{stfloats}
\usepackage{url}
\usepackage{soul}
\usepackage{verbatim}
\usepackage{graphicx}
\usepackage{orcidlink}
\usepackage{cite}
\let\labelindent\relax
\usepackage{enumitem}
\usepackage{marginnote}

\newcommand{\cmark}{{\color{green}\ding{51}}} 
\newcommand{\xmark}{{\color{red}\ding{55}}}   

\title{\LARGE \bf
PPGuide: Steering Diffusion Policies with \\Performance Predictive Guidance
}

\author{Zixing Wang$^1$, Devesh K. Jha$^2$, Ahmed H. Qureshi$^1$, Diego Romeres$^3$
\thanks{$^{1}$Department of Computer Science at Purdue University, IN 47907, USA. {\tt\small \{wang5389, ahqureshi\}@purdue.edu}}%
\thanks{$^{2}$Contribution was conducted while the author was at Mitsubishi Electric Research Laboratories. \tt\small{devesh.dkj@gmail.com}}%
\thanks{$^{3}$Mitsubishi Electric Research Laboratories, Cambridge, MA 02139 USA. {\tt\small romeres@merl.com}}%
}

\let\oldtwocolumn\twocolumn
\renewcommand\twocolumn[1][]{
    \oldtwocolumn[{#1}{
    \begin{center}
            \includegraphics[width=\linewidth]{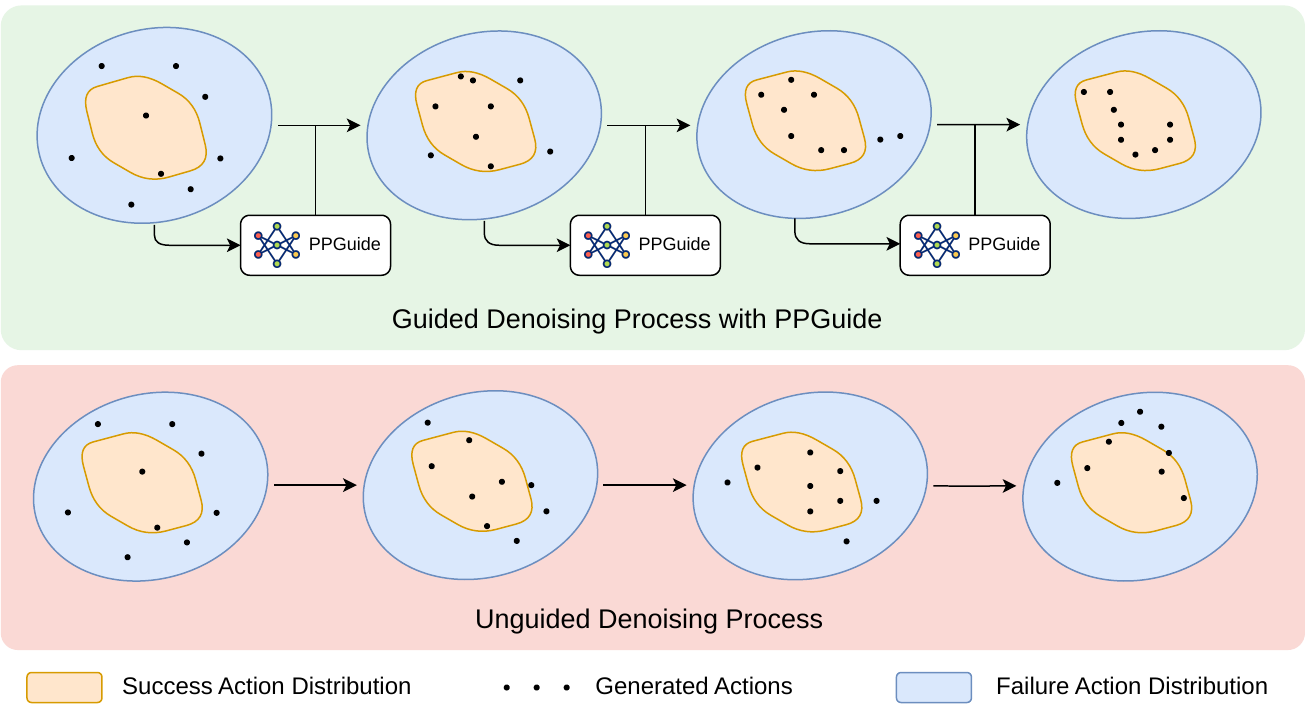}
            \captionof{figure}{PPGuide is a policy steering framework to improve performance of pre-trained diffusion policies at inference time. PPGuide makes use of a learned classifier to guide the pre-trained denoising process, estimating observation-action chunks associated with failure and redirecting them towards the distribution of successful actions, resulting in a more robust policy. It is noted that the schematic shown here is just for presentation purposes.}
            \label{fig:title}
    \end{center}
    }]
}

\begin{document}

\maketitle
\thispagestyle{empty}
\pagestyle{empty}

\begin{abstract}
Diffusion policies have shown to be very efficient at learning complex, multi-modal behaviors for robotic manipulation. However, errors in generated action sequences can compound over time which can potentially lead to failure. Some approaches mitigate this by augmenting datasets with expert demonstrations or learning predictive world models which might be computationally expensive. We introduce Performance Predictive Guidance (PPGuide), a lightweight, classifier-based framework that steers a pre-trained diffusion policy away from failure modes at inference time. PPGuide makes use of a novel self-supervised process: it uses attention-based multiple instance learning to automatically estimate which observation-action chunks from the policy's rollouts are relevant to success or failure. We then train a performance predictor on this self-labeled data. During inference, this predictor provides a real-time gradient to guide the policy toward more robust actions. We validated our proposed PPGuide across a diverse set of tasks from the Robomimic and MimicGen benchmarks, demonstrating consistent improvements in performance.

\end{abstract}


\section{Introduction}


Diffusion policies~\cite{Chi2023} have been shown to be very powerful and efficient at learning complex, long-horizon multi-modal action distributions.
Diffusion policies learn a conditional generative model using demonstration data and they can learn meaningful policies for a large variety of tasks.
However, the stochastic nature of the underlying generative models can lead to compounding errors over time. Over long horizons, subtle errors in generated action chunks can compound, leading to catastrophic drift and task failure. This makes the learned diffusion policy brittle to slight variations during execution. In this paper, we present a guidance method that can steer the pre-trained denoising process to direct them towards more successful actions resulting in robust policies (see Figure~\ref{fig:title}). It is noted that the proposed method does not require access to demonstration data and could be applied to pre-trained policy during inference.

To improve robustness of diffusion policies, existing approaches typically fall into two categories. Data-centric methods~\cite{ross2011reduction, laskey2017dart, Mandlekar2021} rely on dataset augmentation, such as increasing volume, enhancing diversity, or incorporating corrective demonstrations. While effective, these approaches require substantial human effort for data collection and annotation. Reward-based methods leverage dense task rewards when available, either through reinforcement learning fine-tuning~\cite{chen2025fdpp, ren2024diffusion} or residual action learning~\cite{yuan2024policy, ankile2025imitation, haldar2023teach, wang2025implicit}. However, dense rewards are often unavailable or expensive to engineer in real-world scenarios. Recent advances in inference-time guidance~\cite{dhariwal2021diffusion} have opened new possibilities for policy steering. Several works~\cite{Ajay2022, janner2022planning, Mishra2023, nakamoto2025steeringgeneralistsimprovingrobotic, Reuss2023, sun2025latent, wang2025inference, du2025dynaguidesteeringdiffusionpolices} have demonstrated impressive performance gains by guiding diffusion policies at inference time by adjusting the denoising process. However, these methods typically require either dense reward signals~\cite{Ajay2022, janner2022planning} or accurate world models~\cite{Reuss2023, du2025dynaguidesteeringdiffusionpolices}, both of which may be unavailable or computationally prohibitive in practice.

In this work, we present a classifier-based guidance-based framework that improves the robustness of pre-trained diffusion policies using only sparse, binary terminal rewards (e.g., success or failure). The core challenge is to estimate relevance using sparse success or failure signals: how can a sparse, trajectory-level outcome provide dense, actionable guidance for the policy at each timestep. Instead of relying on auxiliary models to estimate state value or out-of-distribution scores~\cite{sun2025latent, du2025dynaguidesteeringdiffusionpolices}, our approach directly learns to estimate which actions within a trajectory are most \emph{relevant} to the final outcome.


We draw inspiration from attention-based Multiple Instance Learning (MIL) in computer vision, where models learn to estimate specific regions of interest from weak image-level labels~\cite{li2015multiple, ilse2018attention, li2021dual}. We hypothesize that a similar principle can estimate success-relevant and failure-relevant state-action subsequences within a robotic trajectory, using only a binary outcome label. Based on this insight, we present \textbf{P}erformance \textbf{P}redictive \textbf{G}uidance (\textbf{PPGuide}). We use a two-stage learning process: first, we use an attention-based MIL model to automatically discover and label which observation-action chunks are most predictive of task outcomes. Second, we train a lightweight relevance classifier on these self-generated labels. During inference, this classifier provides a dense guidance signal, a gradient with respect to the action, steering the diffusion policy away from actions associated with failure. This MIL-based labeling process creates a powerful self-supervised loop, solving the temporal relevance assignment/estimation problem without any manual annotation.

PPGuide offers several advantages: (1)~\textbf{Data-efficient}, it requires only sparse, binary success signals readily available in most robotic tasks; (2)~\textbf{Self-supervised}, it is entirely self-supervised, learning from the policy's own experiences without external supervision; (3)~\textbf{Lightweight}, it adds minimal computational overhead during inference; and (4)~\textbf{Model-agnostic}, applicable to any pre-trained diffusion model based policy without architectural changes.

We validate PPGuide on a diverse suite of challenging manipulation tasks from the Robomimic benchmark~\cite{robomimic2021}. Our results demonstrate that PPGuide substantially improves task success rates over the base diffusion policies, achieving consistent performance gain across different tasks.

\section{Related Work}

\subsection{Steering Robot Control Policies}

Approaches for steering control policies can be broadly categorized by whether they modify the policy's parameters through re-training or gradient-based guidance.

\textbf{Policy Re-training and Fine-tuning.} One major family of approaches involves updating the policy's weights. This includes interactive methods like DAgger~\cite{ross2011reduction, kelly2019hg} and DART~\cite{laskey2017dart}, which collect corrective data from human experts to enrich the training set. While effective, these methods require significant expert involvement. Another popular technique is to use reinforcement learning (RL) to fine-tune a policy's parameters~\cite{ren2024diffusion, chen2025fdpp} or train a residual policy that corrects the base actions~\cite{yuan2024policy, ankile2025imitation, haldar2023teach}. However, RL often requires extensive and potentially unstable training phases.

\textbf{Gradient-based Inference-Time Steering.} To circumvent the costs of re-training, a second family of methods focuses on modifying the policy's output at runtime with gradient-based guidance. The concept of steering a generative model with an auxiliary function has been highly successful, particularly in image synthesis. Classifier guidance~\cite{dhariwal2021diffusion} uses the gradient of a trained classifier $p(y \mid x)$ to steer a generative model $p(x)$ towards producing samples $x$ that belong to a desired class $y$. This idea was central to large-scale diffusion models~\cite{ho2020denoising}. 

Gradient-based techniques include guided denoising, which injects gradient guidance from reward signals directly into the sampling process of diffusion policies~\cite{Ajay2022, janner2022planning, Reuss2023}; value-based filtering, which selects the best action from a set of proposals~\cite{nakamoto2025steeringgeneralistsimprovingrobotic}; and human-in-the-loop steering, where user constraints guide the sampler~\cite{wang2025inference, Wu2025}. An important sub-category is predictive steering, which uses a world or dynamics model to anticipate future outcomes and steer the policy toward safe and robust actions~\cite{sun2025latent, Qi2025, du2025dynaguidesteeringdiffusionpolices, Sun2024, wang2023deri}. PPGuide belongs to the inference-time steering category but introduces a distinct mechanism that requires neither dense rewards nor an explicit world model. The closest related work is Latent Policy Barrier (LPB)~\cite{sun2025latent}, which uses a world model to predict future out-of-distribution observations and steers the policy away from them. In contrast, PPGuide does not need a dynamics model and instead learns to directly assess the relevance of an action for the final task outcome using a self-supervised MIL classifier. This allows it to generate useful guidance gradients without the computational overhead or data requirements of training an auxiliary world model.




\begin{figure*}[t!]
    \vspace{5mm}
    \centering
    \includegraphics[width=\textwidth]{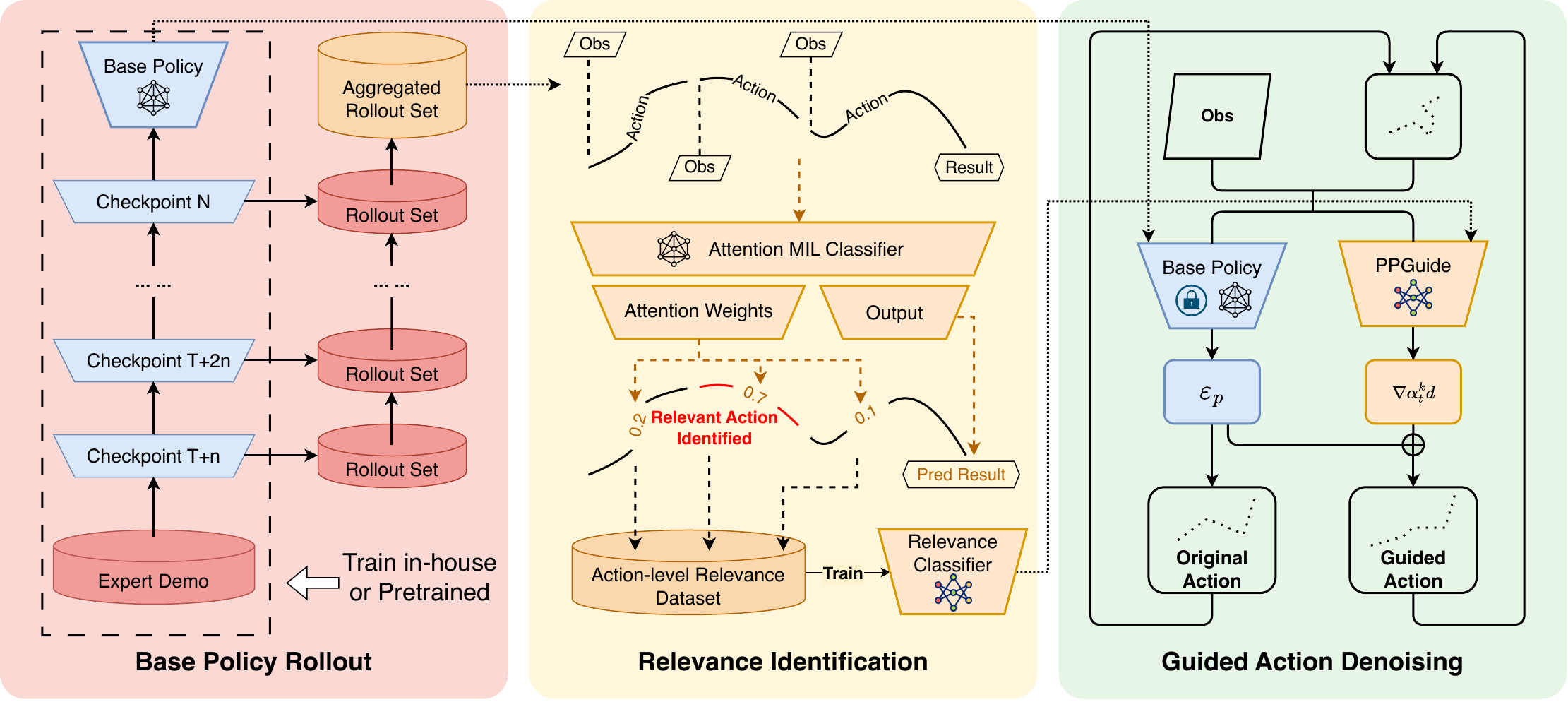}
    \caption{\textbf{Overview of our PPGuide framework.} First, a diverse dataset of trajectories is collected using policy checkpoints from different training stages. Then, (1) an attention-based MIL model analyzes these trajectories to automatically label observation-action chunks as Success-Relevant (SR), Failure-Relevant (FR), or Irrelevant (IR). (2) A lightweight classifier is trained on this labeled data to predict relevance from observation-action pairs. Finally, during inference, gradients from the classifier steer the diffusion sampling process away from failure modes while promoting success-relevant behaviors.}
    \label{fig:pipeline}
    \vspace{-2mm}
\end{figure*}

\subsection{Applications of Multiple Instance Learning}

Multiple Instance Learning (MIL) is a weakly supervised paradigm where labels are applied to bags of instances rather than individual data points~\cite{dietterich1997solving}. While foundational methods introduced concepts like Diverse Density~\cite{NIPS1997_82965d4e}, the field has been transformed by deep learning, particularly through attention mechanisms that learn to weight the importance of each instance~\cite{ilse2018attention}. This modern approach has been especially impactful in computational pathology, where Transformers treat patches from Whole Slide Images as instances to model their relationships for classification~\cite{shao2021transmil}. Subsequent refinements have focused on improving computational efficiency~\cite{li2021dual} and interpretability~\cite{zhang2022dtfd}. The versatility of deep MIL is further demonstrated across diverse applications, from object detection~\cite{li2015multiple}, survival analysis~\cite{yao2020whole} to video anomaly detection~\cite{feng2021mist, lv2023unbiased}. Perhaps the closest work to ours is~\cite{mahmoudieh2020weaklysupervised}, which extract the reusable skills from expert demonstrations to boost imitation policy learning. Our framework, however, introduces a novel two-stage approach that uniquely combines MIL with classifier guidance to steer policy generation. To be best of our knowledge, PPGuide is the first work combining MIL with the guided diffusion denoising process.

\section{Method}
We present a method for improving robustness of pre-trained  diffusion policies. We consider a policy $\pi_\theta$ that, following standard conventions, receives a history of $T_o$ observations $\mathbf{OS}_t^{T_o} = \{o_{t-T_o+1}, \ldots, o_t\}$ and generates a chunk of $T_p$ future actions $\mathbf{AS}_t^{T_p} = \{a_t, \ldots, a_{t+T_p-1}\}$ via an iterative denoising process, after which the first $T_a$ actions are executed. While effective, such policies can still generate action chunks that lead to task failure. The fundamental challenge is to identify and correct these failure-prone actions online, without access to fine-grained manual labels; the only available supervision is typically the final outcome of a long-horizon trajectory.

To address this, we introduce \textbf{PPGuide}, a framework for classifier-guided policy steering that operates in three phases, as shown in Figure~\ref{fig:pipeline}. First, we begin with pre-trained diffusion policy, which we term the base policy. Our framework is agnostic to the origin of this model; it can be a policy trained or a publicly available, pre-trained model accessed directly. Then, we build a dataset by collecting the rollouts from checkpoints at different training stages. Next, in an offline relevance estimation or identification phase, we frame the problem of estimating/identifying actions relevant to the task result through the lens of Multiple Instance Learning (MIL). We automatically analyze a diverse dataset of successful and failed trajectories to localize the specific observation-action chunks which are most likely responsible for the outcome. Then, using these localized observation-actions chunks as pseudo-labels, we train a lightweight guidance classifier to predict the relevance of any given observation-action chunk in real-time. Finally, during inference, we employ this classifier to steer the policy's denoising process. By using gradients from the classifier, we actively guide the sampling towards successful behaviors and away from predicted failure modes, enhancing overall task success and robustness.

\subsection{Offline Estimation of Relevant Actions}

\subsubsection{Solving Problem with Multiple Instance Learning}
A core challenge in refining diffusion policies lies in estimating which specific actions within a long-horizon trajectory are relevant to the eventual outcome. Manually annotating every observation-action chunk as success-relevant, failure-relevant, or irrelevant is prohibitively expensive and often ambiguous. The only readily available supervisory signal is the final outcome of the entire trajectory (i.e., task success or failure). We found this scenario, where a single label / sparse reward is assigned to a collection of instances, is a natural fit for the Multiple Instance Learning (MIL) paradigm~\cite{dietterich1997solving}.

In our framework, we formulate this relevance estimation/identification problem as a binary MIL task. A complete trajectory, $\mathcal{T} = \{(os_0^j, as_0^k), (os_1^j, as_1^k), \dots, (os_{N-1}^j, as_{N-1}^k)\}$, consisting of a sequence of observation-action chunk pairs, is treated as a \textbf{bag}. Each individual action chunk of length $k$ at step $t$, $as_t^k$, conditioned on its corresponding observation $os_t^j$ chunk of length $j$ at step $t$, is an~\textbf{instance} within that bag. The trajectory is assigned a bag-level label, $Y \in \{\text{success, failure}\}$.

The standard MIL assumption posits that a positive bag contains at least one "witness" or positive instance, while a negative bag is composed entirely of negative instances. Our problem deviates slightly from this classic definition~\cite{dietterich1997solving}, adopting a more generalized but equally valid formulation. Rather than a simple presence-versus-absence model, our task involves distinguishing between bags containing one of two distinct types of relevant instances. We define our learning objective as follows:
\begin{itemize}
    \item A success bag (i.e. $T$) contains at least one \textbf{success-relevant instance} (i.e. $(os^j, as^k)$), an observation-action chunk that decisively contributes to achieving the task goal.
    \item A failure bag contains at least one \textbf{failure-relevant instance}, an observation-action chunk that leads to an irrecoverable or terminal error state.
\end{itemize}

Under this formulation, the MIL model is not learning to detect "something vs. nothing," but rather to estimate the presence of "evidence of success" versus "evidence of failure" within the bag. The objective of our attention-based MIL classifier, is to predict the bag label $Y$ given the set of instances $\{as_t\}_{t=0}^{N-1}$ in $\mathcal{T}$. Crucially, the attention mechanism is trained to assign high weights to the instances most indicative of the bag's label, effectively localizing the success-relevant and failure-relevant observation-action chunks without requiring explicit instance-level supervision. This process allows us to generate a pseudo-labeled dataset for training a subsequent instance-level classifier for online policy guidance. To ensure a diverse dataset that captures a wide range of policy behaviors, from premature to proficient, we collect trajectories by rolling out checkpoints of the diffusion policy $\pi_\theta$ at various stages of its training.

\subsubsection{Instance-Level Relevance Estimation via Attention}

To implement our MIL classifier, we employ a neural network with a gated attention mechanism~\cite{ilse2018attention}. First, each instance $(os_t^j, as_t^k)$ within a trajectory bag is passed through an instance encoder, $\phi$, to generate a low-dimensional embedding $h_t = \phi(os_t^j, as_t^k)$. This encoder is typically a multi-layer perceptron (MLP) that maps the concatenated observation and action features into a shared embedding space.

Given the sequence of instance embeddings $H = \{h_0, h_1, \dots, h_{N-1}\}$, the attention mechanism computes a weight $\alpha_t$ for each instance, signifying its contribution to the final bag-level prediction. The attention weights are calculated as:
\begin{equation}
    \alpha_t = \frac{\exp\left(w^\top \left( \tanh(V h_t^\top) \odot \text{sigm}(U h_t^\top) \right) \right)}{\sum_{j=0}^{N-1} \exp\left(w^\top \left( \tanh(V h_j^\top) \odot \text{sigm}(U h_j^\top) \right) \right)}
\end{equation}
where $V$, $U$, and $w$ are learnable weight matrices of the attention network, $\odot$ denotes element-wise multiplication, and $\text{sigm}$ is the sigmoid function. This gated attention formulation provides additional representative power over a standard softmax.

The trajectory-level feature representation, $z$, is then formed by a weighted sum of the instance embeddings:
\begin{equation}
    z = \sum_{t=0}^{N-1} \alpha_t h_t
\end{equation}
Finally, a classifier $g$ maps this aggregated representation to a prediction for the bag label, $P(Y | \mathcal{T}) = g(z)$. The entire model, including the encoder $\phi$, the attention network, and the classifier $g$, is trained end-to-end using a binary cross-entropy loss against the true trajectory labels.


\subsection{Online Instance-Level Guidance Classifier}

\subsubsection{Constructing the Labeled Instance Dataset via MIL}
After training the MIL model, we use it to generate our instance-level dataset, $\mathcal{D}_{inst}$. We perform a forward pass with the trained model on our complete set of rollout trajectories. For each trajectory $\mathcal{T}$, we compute the attention weights $\{\alpha_t\}_{t=0}^{N-1}$ for all its observation-action chunks.

An observation-action chunk $(os_t^j,~as_t^k)$ is deemed "relevant" if its attention weight exceeds a predefined threshold $\tau$. In this work, we use z-score, i.e., standard score, to divide the chunks. We partition the instances into three distinct classes based on the weights and the trajectory's outcome:
\begin{itemize}
    \item \textbf{Success-Relevant (SR):} Instances from successful trajectories where $\alpha_t > \tau$. \\
    $\mathcal{D}_{SR} = \{(os_t^j,~as_t^k) \mid Y_\mathcal{T}=\text{success} \land \alpha_t > \tau \}$
    \item \textbf{Failure-Relevant (FR):} Instances from failed trajectories where $\alpha_t > \tau$. \\
    $\mathcal{D}_{FR} = \{(os_t^j,~as_t^k) \mid Y_\mathcal{T}=\text{failure} \land \alpha_t > \tau \}$
    \item \textbf{Irrelevant (IR):} All other instances where the attention weight is below the threshold, regardless of trajectory outcome. \\
    $\mathcal{D}_{IR} = \{(os_t^j,~as_t^k) \mid \alpha_t \leq \tau \}$
\end{itemize}
The final dataset is the union of these three sets, $\mathcal{D}_{inst} = \mathcal{D}_{SR} \cup \mathcal{D}_{FR} \cup \mathcal{D}_{IR}$. This dataset forms the basis for training the supervised classifier used for policy guidance. During implementation, we observed that observation-action chunks classified as IR outnumbered those deemed SR or FR by more than tenfold. This incidentally highlights the discriminative power of our MIL model to pinpoint the few critical moments within a long trajectory.

With the pseudo-labeled instance dataset $\mathcal{D}_{inst}$ established, we train a standard supervised classifier, $f_{guide}$, to act as an oracle for online guidance. This classifier is modeled as a neural network that takes an observation-action pair $(os_t^j,~as_t^k)$ as input and outputs a probability distribution over the three instance-level classes, $P_{f_{guide}}(y | os_t^j,~as_t^k)$, where $y \in \{\text{SR, FR, IR}\}$. The model is trained using a standard multi-class cross-entropy loss. The primary function of $f_{guide}$ works at inference time by providing a gradient signal that quantifies whether an action is likely to lead to success or failure, enabling proactive correction.


\begin{figure*}[t!]
 \vspace{5mm}
    \centering
    \captionsetup{justification=centering}
    \includegraphics[width=\textwidth]{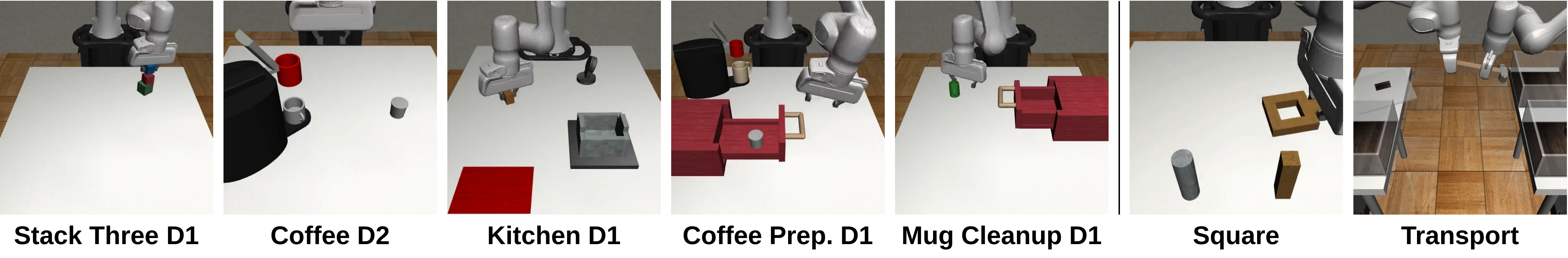}
    \caption{Evaluation tasks from the Robomimic and MimicGen benchmarks. (The \textbf{Stack D1} task uses two cubes, while \textbf{Stack Three D1} uses three).}
    \label{fig:tasks}
    \vspace{2mm}
\end{figure*}

\begin{table*}[t]
\vspace{-2mm}
\setlength{\tabcolsep}{4pt}
\renewcommand{\arraystretch}{1.2}
    \begin{center}
        \vspace{-2mm}
        \begin{tabular}{c | c | c | c | c | c | c | c | c }
        \toprule
        Challenge        & Stack D1 & Stack Three D1 & Coffee D2 & Coffee Prep. D1 & Kitchen D1 & Mug Cleanup D1 & Square & Transport \\ \midrule
        Long-horizon     & \xmark   & \xmark         & \xmark    & \cmark          & \cmark     & \cmark         & \xmark & \cmark  \\
        Precision        & \xmark   & \xmark         & \cmark    & \cmark          & \xmark     & \xmark         & \cmark & \xmark  \\
        Articulated Obj. & \xmark   & \xmark         & \cmark    & \cmark          & \xmark     & \cmark         & \xmark & \xmark \\
        \bottomrule
        \end{tabular}
     \end{center}
     \caption{Task specifications across different benchmark tasks.}
     \label{tb:tasks}
    \vspace{-3mm}
\end{table*}

\subsection{Alternating Classifier Guidance for Policy Refinement}

During inference, we modify the standard reverse diffusion process of the policy to incorporate guidance from the trained classifier $f_{guide}$. The diffusion policy generates an action chunk $as_t$ by starting with Gaussian noise $as_t^K \sim \mathcal{N}(0, I)$ and iteratively denoising it for $K$ steps. At each denoising step $k \in \{K, \dots, 1\}$, the model predicts the noise $\epsilon_\theta(as_t^k, k, os_t^j)$ that was added to the clean action.

To refine the policy's behavior, we steer the denoising process to simultaneously encourage SR actions and discourage FR actions. This is achieved using gradients from the guidance classifier's log-probabilities with respect to the action at step $k$:
\begin{align}
g_{sr}(as_t^k,os_t^j) &= \nabla_{as_t^k} \log P_{f_{guide}}(y=\text{SR} |os_t^j,as_t^k) \\
g_{fr}(as_t^k,os_t^j) &= \nabla_{as_t^k} \log P_{f_{guide}}(y=\text{FR} |os_t^j,as_t^k)
\end{align}
The gradient $g_{sc}$ points in the direction that increases the likelihood of being success-relevant, while $g_{fc}$ does the same for failure-relevant. We combine these signals to create a modified noise estimate, $\hat{\epsilon}_\theta$:
\begin{equation}
\begin{split}
\hat{\epsilon}_\theta(as_t^k,k,os_t^j) = &~\epsilon_\theta(as_t^k, k, os_t^j) + \\
& w_{sr} \cdot g_{sc}(as_t^k, os_t^j) - w_{fc} \cdot g_{fr}(as_t^k, os_t^j)
\end{split}
\end{equation}
Here, $w_{sr}$ and $w_{fr}$ are two separate scalar hyperparameters that control the strength of the attraction towards success and the repulsion from failure, respectively. It is worth noting that $w_{sr}$ shall be much lower than $w_{fr}$ for good performance. This asymmetry is motivated by the differing nature of success and failure in manipulation tasks. Failure modes are often diverse and can occur at many points, making a strong, general repulsion from FR patterns a robust strategy. Conversely, SR actions are typically sparse and context-specific (e.g., the final grasp). A strong, constant attraction towards these SR patterns can destabilize the policy by forcing IR parts of the trajectory to conform to a pattern that is only relevant at a specific moment.

Crucially, although our classifier is lightweight, it add computational overheads due to the iterative denoising process. Instead of applying the guidance to every step, we employ an alternating guidance schedule, applying the correction only on, for example, even-numbered denoising steps. According to the experiments, we found this strategy can achieve almost the same performance as constant guidance while reduce significant network forwarding computation.

\begin{table*}[t]
\vspace{6mm}
\centering
\setlength{\tabcolsep}{6pt} 
\renewcommand{\arraystretch}{1.35}
\begin{center}
\vspace{-3mm}

\begin{tabular}{c  cc  cc  cc  cc}
\toprule
Method 
    & \multicolumn{2}{c}{Stack D1} 
    & \multicolumn{2}{c}{Stack Three D1} 
    & \multicolumn{2}{c}{Coffee D2} 
    & \multicolumn{2}{c}{Coffee Prep. D1} \\ 
\cmidrule(lr){2-3}\cmidrule(lr){4-5}\cmidrule(lr){6-7}\cmidrule(lr){8-9}
Policy Epochs
    & 500 & 550
    & 500 & 550
    & 500 & 550
    & 500 & 550 \\ \midrule
    
DP                 & 92\%                 & 92\%                  & 28\%                  & 30\%                  & 54\%                  & 46\%                    & 16\%                  & 18\%        \\
DP-SS              & 88\% (-4\%)          & 90\% (-2\%)           & 24\% (-4\%)           & 26\% (-4\%)           & 44\% (-10\%)          & \textbf{60\% (+14\%)}   & 14\% (-2\%)           & 14\% (-4\%) \\
PPGuide-SS         & 92\% (+0\%)          & 90\% (-2\%)           & 32\% (+4\%)           & \textbf{34\% (+4\%)}  & 56\% (+2\%)           & \textbf{60\% (+14\%) }  & \textbf{24\% (+8\%) } & 20\% (+2\%) \\
PPGuide-CG         & \textbf{94\% (+2\%)} & \textbf{94\% (+2\%)}  & \textbf{36\% (+8\%)}  & 32\% (+2\%)           & \textbf{58\% (+4\%)}  & 58\% (+12\%)            & 20\% (+4\%)           & \textbf{24\% (+6\%)} \\
PPGuide          & \textbf{94\% (+2\%)} & \textbf{94\% (+2\%)}  & 34\% (+6\%)           & 28\% (-2\%)           & \textbf{58\% (+4\%)}  & 58\% (+12\%)            & 20\% (+4\%)           & 22\% (+4\%) \\
\bottomrule
\end{tabular}

\vspace{2mm}
    
\begin{tabular}{c  cc  cc  cc  cc}
\toprule
Method 
& \multicolumn{2}{c}{Kitchen D1} 
& \multicolumn{2}{c}{Mug Cleanup D1} 
& \multicolumn{2}{c}{Square} 
& \multicolumn{2}{c}{Transport} \\ 
\cmidrule(lr){2-3}\cmidrule(lr){4-5}\cmidrule(lr){6-7}\cmidrule(lr){8-9}
Policy Epochs
    & 500 & 550
    & 500 & 550
    & 500 & 550
    & 500 & 550 \\ \midrule
        
DP                 & 52\%                  & 40\%                  & 26\%                  & 26\%                      & 62\%                   & 58\%                   & 60\%                    & 68\%        \\
DP-SS              & 38\% (-14\%)          & 36\% (-4\%)           & 24\% (-2\%)           & 34\% (+8\%)               & 58\% (-4\%)            & 56\% (-2\%)            & 54\% (-6\%)             & 58\% (-10\%) \\
PPGuide-SS         & 44\% (-8\%)           & 38\% (-2\%)           & \textbf{34\% (+8\%)}  & 32\% (+6\%)               & 68\% (+6\%)            & 60\% (+8\%)            & 62\% (+2\%)             & 62\% (-6\%) \\
PPGuide-CG         & \textbf{54\% (+2\%)}  & \textbf{44\% (+4\%)}  & 32\% (+6\%)           & 32\% (+6\%)               & \textbf{72\% (+10\%)}  & \textbf{68\% (+10\%)}  & \textbf{68\% (+8\%)}    & 74\% (+6\%) \\
PPGuide            & 52\% (+0\%)           & \textbf{44\% (+4\%)}  & 30\% (+4\%)           & \textbf{36\% (+10\%)}     & \textbf{72\% (+10\%)}  & 66\% (+8\%)            & \textbf{68\% (+8\%)}    & \textbf{76\% (+8\%)} \\ \bottomrule 
\end{tabular}
        
\end{center}
\vspace{0mm}
\caption{Benchmark Evaluation Results.}
\label{tb:results}
\end{table*}


\begin{table*}[t]
\vspace{2mm}
\centering
\setlength{\tabcolsep}{6pt} 
\renewcommand{\arraystretch}{1.35}
\begin{center}
\vspace{-3mm}

\begin{tabular}{c  cccc  cccc}
\toprule
Method 
    & \multicolumn{4}{c}{Square} 
    & \multicolumn{4}{c}{Transport} \\ 
\cmidrule(lr){2-5}\cmidrule(lr){6-9}
Policy Epochs
    & 1300 & 1400 & 1500 & 1600 & 1300 & 1400 & 1500 & 1600 \\ \midrule

DP                       & 54\%                   & 60\%         & 62\%                 & 62\%                     & 56\%                  & 68\%                 & \textbf{74\%}   & 58\%        \\
\textbf{PPGuide (Ours)}  &~\textbf{70\% (+16\%)}  & 60\% (+0\%)  & \textbf{68\% (+6\%)} & \textbf{70\% (+8\%)}     & \textbf{74\% (+18\%)} & \textbf{72\% (+4\%)} & 70\% (-4\%)     & \textbf{70\% (+12\%)} \\
\bottomrule
\end{tabular}
\end{center}
\vspace{0mm}
\caption{Heterogeneous Evaluation Results.}
\label{tb:hetero_results}
\end{table*}

\section{Experiments}

\subsection{Simulation Environment and Dataset}
Our experiments are conducted on various tasks from Robomimic~\cite{robomimic2021} and Mimicgen~\cite{mandlekar2023mimicgen}, two large-scale environments for robotic imitation learning. Our experiment tasks include long-horizon and articulated objects manipulation, as presented in Table.~\ref{tb:tasks} and Figure~\ref{fig:tasks}. These benchmarks also provide tele-operated and synthesized expert demonstrations, which we use to train the base policies. To assess the sample efficiency of our method, we simulate a limited-data scenario by training the base policies on only a 10\% subset of the original expert demonstrations for each task. To train PPGuide, we collect rollout data from a series of checkpoints saved during this training process, specifically at epochs 250, 300, 350, 400, and 450. For the final evaluation, we then apply the resulting PPGuide at inference time to guide two distinct, later-stage checkpoints (epochs 500 and 550). This setup allows us to test PPGuide's ability to improve performance in a low-data regime and its generalization across different base policy checkpoints.

We compare PPGuide with the following baselines:
\begin{itemize}
    \item \textbf{DP}: Diffusion Policy~\cite{Chi2023}, which also serve as the base policy of PPGuide.
    \item \textbf{DP-SS}: Diffusion Policy~\cite{Chi2023} with \textbf{s}tochastic \textbf{s}ampling, which helps stabilizing the denoising process via Markov chain Monte Carlo~\cite{wang2025inference}.
    \item \textbf{PPGuide-CG}: PPGuide with \textbf{c}onstant \textbf{g}uidance of denoising process (thus guidance is provided at every denoising step). The strength is the same as PPGuide for each task.
    \item \textbf{PPGuide-SS}: PPGuide with \textbf{s}tochastic \textbf{s}ampling for denoising process guidance. We use 4 sampling steps as ITPS~\cite{wang2025inference} and DynaGuide~\cite{du2025dynaguidesteeringdiffusionpolices} did. 
    \item \textbf{PPGuide}: PPGuide with alternating guidance schedule, balancing the performance and inference speed.
    
\end{itemize}

As PPGuide and its baseline variants involves several key hyperparameters, we report the best results obtained from a limited grid search. We note that this search was not exhaustive, and these results may not represent the upper bound of the method's performance. 

\subsection{Benchmark Evaluation Results and Analysis}
The evaluation results in Table~\ref{tb:results} show that PPGuide consistently matches or exceeds all baselines in the limited-demonstration setting, indicating strong sample efficiency. The performance gains are most substantial on the long-horizon and precision-sensitive tasks, highlighting our method's effectiveness at mitigating the compounding errors that arise from small action deviations over time.

\begin{figure}[t!]
    \centering
    \includegraphics[width=0.48\textwidth]{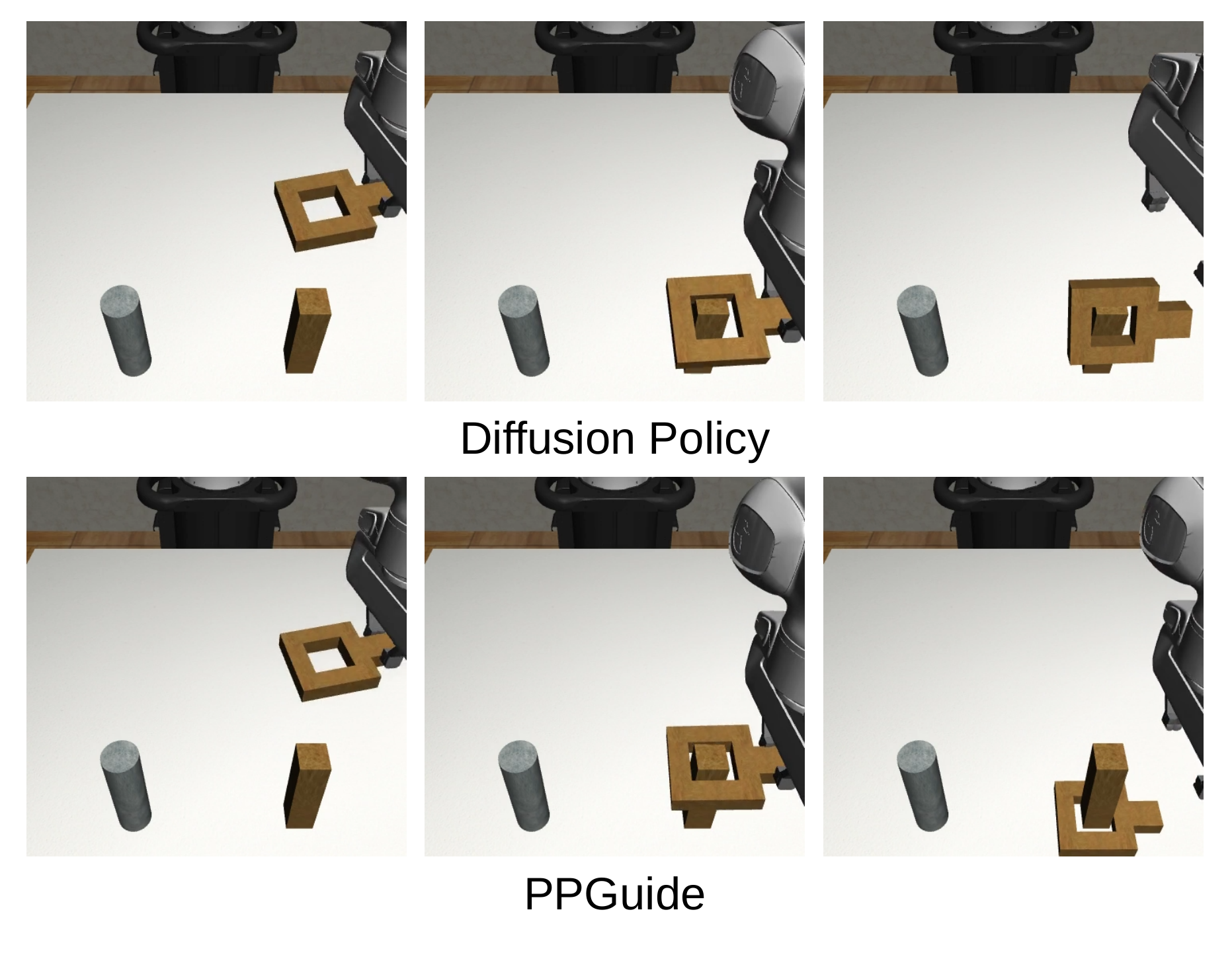}
    \caption{This example shows how PPGuide steers the base policy to avoid misalignment during square insertion.}
    \label{fig:tasks}
    \vspace{-2mm}
\end{figure}


\textbf{Benefit of the Alternating Guided Denoising Process.} As shown in Table~\ref{tb:results}, PPGuide's performance achieved almost the same level of performance as PPGuide-CG with much less guidance inference, which is more suitable for real-time inference. PPGuide-CG and PPGuide consistently outperforms the stochastic sampling (PPGuide-SS) variants. We conclude stochastic sampling is not suitable for PPGuide framework. We do not have a solid assumption on the cause of this result, so we only provide the experiment results for reference.~\textit{We note that this conclusion is specific to our performance-predictive guidance method and the experimental tasks evaluated}.

\textbf{Heterogeneous Base Policies Performance.} In practice, the policy used for data collection (rollout) may differ from the one used for deployment. We evaluated PPGuide's robustness in this heterogeneous setting by training it on rollouts from a series of policies trained for 250, 300, 350, 400 and 450 epochs. We then used this single PPGuide instance to guide a separate series of more extensively trained deployment policies (1300, 1400, 1500, 1600 epochs). The results in Table~\ref{tb:hetero_results} show a significant performance increase across the board. Notably, the improvement on the 1300-epoch checkpoint was among the highest recorded in this paper, underscoring PPGuide's ability to learn a robust and transferable guidance model that is not overfitted to the specific weights of the rollout policy.

\textbf{Sensitivity to Guidance Strength.} As with other guidance-based methods~\cite{sun2025latent, du2025dynaguidesteeringdiffusionpolices, wang2025inference}, the performance of PPGuide depends on the guidance strength hyperparameter, which balances adherence to the guidance signal against fidelity to the learned data distribution. By varying this value while keeping all other settings fixed, we observed an expected trade-off: increasing the guidance strength improves performance up to a certain point, after which it can degrade sample quality and cause instability. This confirms that, like other gradient-based guidance methods, proper tuning of this hyperparameter is necessary for optimal results.

\begin{figure}[t!]
    \centering
    \includegraphics[width=0.48\textwidth]{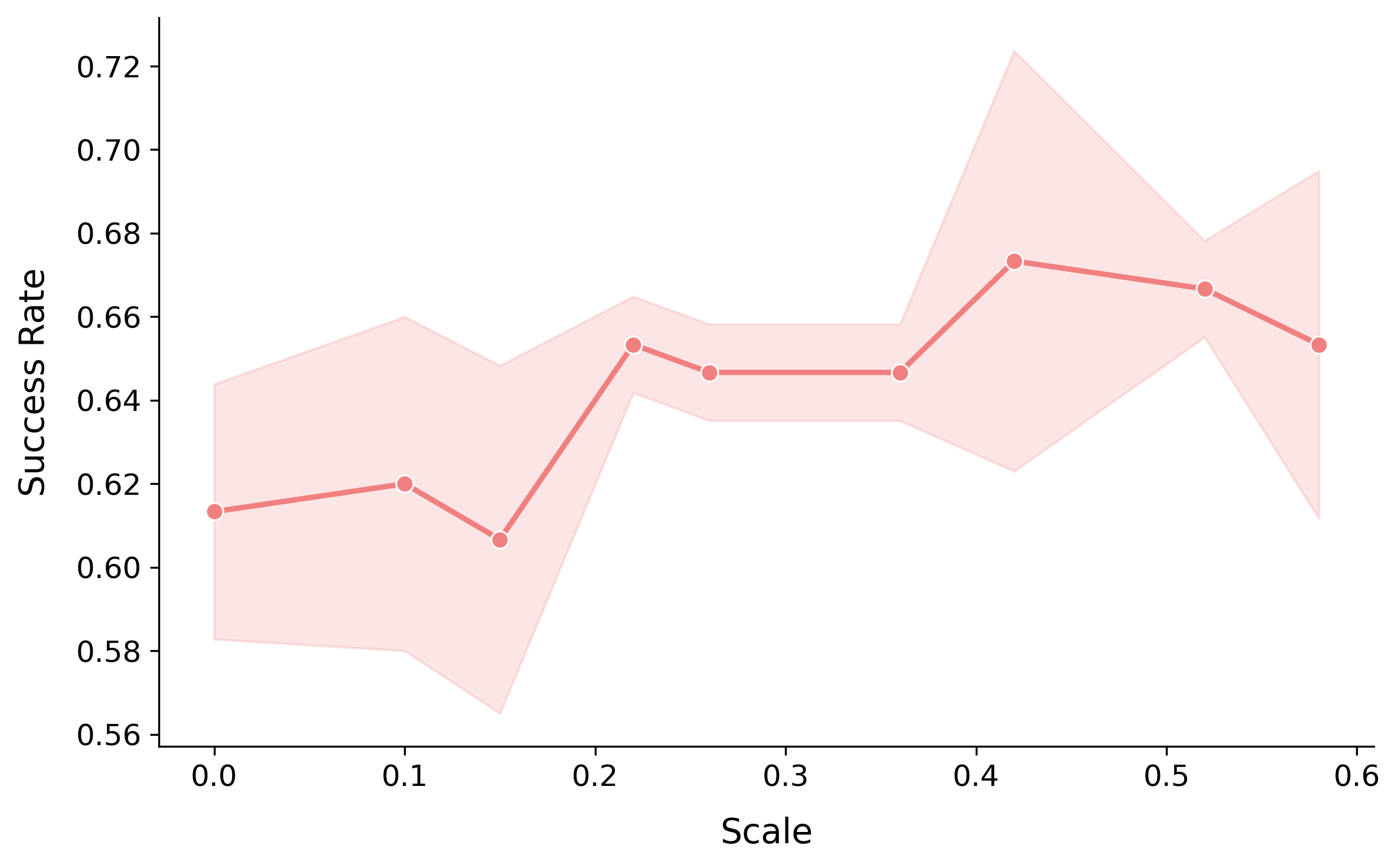}
    \caption{Averaged performance of PPGuide, with base policies of epoch 500 and 550, on the Square task across different guidance strength values. We are able to achieve performance gains over the base DP across a range of guidance strength.}
    \label{fig:mil-id}
    \vspace{-2mm}
\end{figure}

\begin{figure}[t!]
    \centering
    \includegraphics[width=0.48\textwidth]{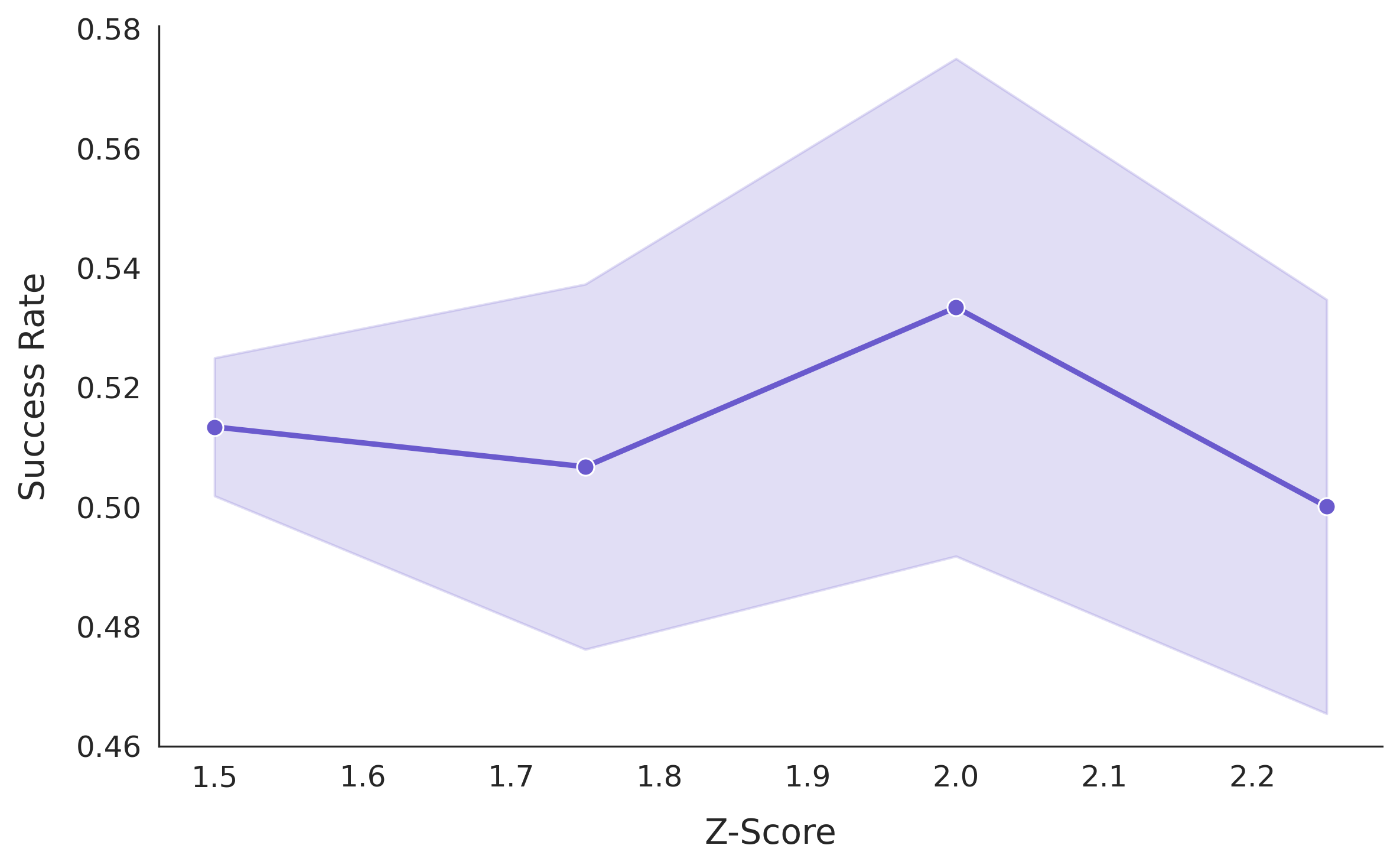}
    \caption{Effect of Z-score selection. We build four datasets to train the classifier using different z-scores for Coffee D2 task. The results are the averaged results of experiments with guidance strength of 0.25, 0.3 and 0.35.
    }
    \label{fig:z-score}
    \vspace{-2mm}
\end{figure}

\textbf{Effect of Dataset Z-Score.} The MIL classifier-created dataset is based on attention weights, where a z-score is used to divide the observation-action chunks into relevant and irrelevant sets, controlling the purity of the self-labeled data.  To analyze this effect, we trained PPGuide's classifier with four datasets created by different z-score thresholds, 1.5, 1.75, 2.0, 2.25, where 2.0 is used across the aforementioned evaluation. The results, plotted in Figure~\ref{fig:z-score}, show that performance peaks at a moderate threshold, 2.0. It also indicates the performance of PPGuide is sensitive to z-score selection, which is a improvement direction in the future work.

\section{Limitations and Future Work}

Despite its performance improvements, our approach has several key limitations rooted in its design. First, its success is fundamentally dependent on the quality of the initial rollouts, as a policy that rarely succeeds presents a ``cold start" problem for our self-labeling process. This process is also susceptible to learning spurious correlations from the initial data, where an irrelevant but recurring feature could be misinterpreted as relevant, leading to flawed guidance. Finally, the practical application of PPGuide is sensitive to key hyperparameters, such as the z-score threshold and guidance strength, which require careful, task-specific tuning to achieve optimal performance.

These limitations suggest several promising directions for future research. To address the data-dependency issues, PPGuide could be integrated with more robust exploration strategies to ensure a diverse and informative set of initial trajectories. Another avenue is to move beyond the current offline training paradigm by developing methods to update the relevance classifier online as the policy gathers new experience, enabling continuous adaptation to environmental shifts. Finally, exploring more sophisticated credit assignment models could extend our approach to tasks where failure results from the slow accumulation of errors rather than discrete, identifiable events.

\section{Conclusions}
In this paper, we proposed PPGuide, a framework for improving the performance and robustness of pre-trained diffusion policies. Our method addresses the critical challenge of temporal credit assignment from sparse rewards by drawing inspiration from Multiple Instance Learning. PPGuide uses self-supervision and low computational overhead, which makes it suitable for deployment. Our extensive experiments on challenging manipulation tasks demonstrate that PPGuide yields substantial improvements in success rates over baseline diffusion policies. Crucially, these gains are achieved without any additional expert demonstrations, dense reward engineering, or auxiliary world models, highlighting the practicality and data-efficiency of our approach.

\balance

\bibliographystyle{IEEEtran}
\bibliography{references}

\end{document}